\documentclass{article}

    \usepackage[preprint, nonatbib]{neurips_2022}

    \usepackage[nonatbib]{neurips_2022}

\usepackage[utf8]{inputenc} 
\usepackage[T1]{fontenc}    
\usepackage{url}            
\usepackage{booktabs}       
\usepackage{amsfonts}       
\usepackage{nicefrac}       
\usepackage{microtype}      
\usepackage{xcolor}         
\usepackage[colorlinks = true,
            linkcolor = royalazure,
            urlcolor  = royalazure,
            citecolor = royalazure,
            anchorcolor = royalazure]{hyperref}
\usepackage{wrapfig}
\usepackage{caption}
\usepackage{csquotes}
\usepackage{seqsplit}
\usepackage[sort&compress, numbers]{natbib}
\usepackage{url}            %
\usepackage{floatrow} %
\usepackage{multirow}
\usepackage{adjustbox}
\usepackage{amsmath}
\usepackage{arydshln}
\usepackage{float}
\floatstyle{plaintop}
\restylefloat{table}
\usepackage{subcaption}
\usepackage{amsmath}
\usepackage{amsthm}
\usepackage{amssymb}
\usepackage{multirow}
\usepackage{bm}

\newtheorem{theorem}{Theorem}
\newtheorem{lemma}[theorem]{Lemma}

\newtheorem{proposition}[theorem]{Proposition}

\newcommand \eps \varepsilon
\renewcommand \epsilon \varepsilon

\usepackage{tikz}
\usetikzlibrary{shapes}
\usetikzlibrary{positioning}
\usetikzlibrary{plotmarks}
\usetikzlibrary{patterns}
\usetikzlibrary{intersections,shapes.arrows}
\usepackage{pgfplots}
\usetikzlibrary{pgfplots.fillbetween}
\definecolor{cadmiumgreen}{rgb}{0.0, 0.42, 0.24}
\definecolor{oldmauve}{rgb}{0.4, 0.19, 0.28}
\definecolor{royalazure}{rgb}{0.0, 0.22, 0.66}
\definecolor{harvardcrimson}{rgb}{0.79, 0.0, 0.09}
\definecolor{lightmauve}{rgb}{0.86, 0.82, 1.0}
\definecolor{darkbrown}{rgb}{0.4, 0.26, 0.13}

\usepackage{enumitem}
\usepackage{tcolorbox}

\definecolor{azure}{rgb}{0.0, 0.5, 1.0}
\tcbuselibrary{skins}
\tcbset{enhanced}

\title{Conditional Contrastive Learning for Improving Fairness in Self-Supervised Learning}
\author{Martin Q. Ma\textsuperscript{\rm 1}, Yao-Hung Hubert Tsai\textsuperscript{\rm 1},  Paul Pu Liang\textsuperscript{\rm 1}, Han Zhao\textsuperscript{\rm 2}, \\
{\bf Kun Zhang\textsuperscript{\rm 1,3}, Ruslan Salakhutdinov\textsuperscript{\rm 1}, \&  Louis-Philippe Morency\textsuperscript{\rm 1}}  \\
\textsuperscript{\rm 1}Carnegie Mellon University \textsuperscript{\rm 2}University of Illinois at Urbana-Champaign\\
\textsuperscript{\rm 3}Mohamed bin Zayed University of Artificial Intelligence\\
\texttt{\{qianlim, yaohungt, pliang, kunz1, rsalakhu, morency\}@cs.cmu.edu} \\
\texttt{hanzhao@illinois.edu} \\
}

\begin{document}
\maketitle

\begin{abstract}
Contrastive self-supervised learning (SSL) learns an embedding space that maps similar data pairs closer and dissimilar data pairs farther apart. Despite its success, one issue has been overlooked: the fairness aspect of representations learned using contrastive SSL. Without mitigation, contrastive SSL techniques can incorporate sensitive information such as gender or race and cause potentially unfair predictions on downstream tasks. In this paper, we propose a Conditional Contrastive Learning (CCL) approach to improve the fairness of contrastive SSL methods. Our approach samples positive and negative pairs from distributions conditioning on the sensitive attribute, or empirically speaking, sampling positive and negative pairs from the same gender or the same race. We show that our approach provably maximizes the conditional mutual information between the learned representations of the positive pairs, and reduces the effect of the sensitive attribute by taking it as the conditional variable.
On seven fairness and vision datasets, we empirically demonstrate that the proposed approach achieves state-of-the-art downstream performances compared to unsupervised baselines and significantly improves the fairness of contrastive SSL models on multiple fairness metrics.
\end{abstract}

\section{Introduction}

Self-supervised learning (SSL) \citep{jing2020self}, especially contrastive self-supervised learning (contrastive SSL) \citep{chen2020simple, he2020momentum}, have performed well in a variety of different vision or language tasks \citep{chen2021empirical, chen2020big, radford2021learning}. Contrastive SSL frameworks \citep{chen2020simple, he2020momentum} first perform a \textit{contrastive pre-training} by pulling together related data pairs (termed \textit{positive pairs}) and pushing away unrelated pairs (termed \textit{negative pairs}) in the embedding space, and then evaluate the learned representation by a \textit{supervised fine-tuning} with labels. 

\begin{figure}
\centering
\includegraphics[width=\linewidth]{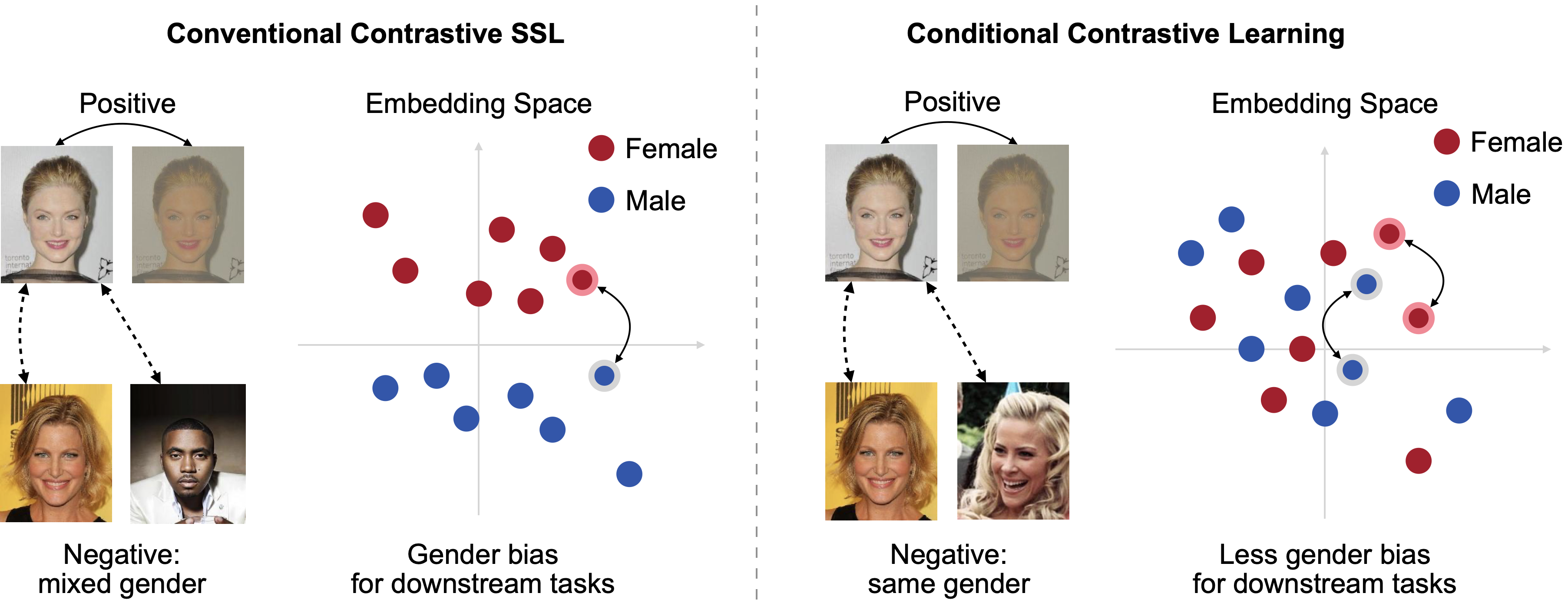}

\caption{A demonstration of conventional contrastive SSL vs. the proposed Conditional Contrastive Learning (CCL). In conventional contrastive SSL, contrastive pre-training is performed on a mixture of female and male samples, and the model can easily pickup gender information during training and create an embedding space with gender bias for downstream tasks. On the other hand, the proposed Conditional Contrastive Learning only samples positive and negative pairs within the same gender group, making it harder for the model to leverage gender information. It can therefore create an embedding space with less gender bias. An empirical reflection of this demonstration on the CelebA dataset \citep{liu2015deep} is shown in Figure \ref{fig:ablation_vision}.}
\label{fig:intro_illu}
\end{figure}

However, despite the growing popularity of contrastive SSL, the potential issue of fairness in these learned representations has been understudied: \textbf{do contrastive SSL models learn fair representations, and how to mitigate potential biases?} We are particularly interested in the scenario where a potential \textit{sensitive attribute}, such as gender or race, is already available in the dataset. We show that without care, contrastive models can incorporate information from the sensitive attributes and cause unfair predictions in downstream tasks. For example, Figure \ref{fig:intro_illu} illustrates a case where contrastive SSL is used to learn representations of human faces. A predominant contrastive SSL setup \citep{chen2020simple} uses two augmented views of the same image as a positive pair, and selects random images as negative pairs. In this setup, two images for the positive pair will always share the same gender since both are augmented from the same image, but two images for the negative pairs may have different genders. Therefore, a contrastive SSL model can learn to use gender-related visual attributes to push away mixed-gender images present in negative pairs. By capturing this gender-related information in the embedding space of the learned representation, the representation can potentially causes unfair predictions when it is applied to downstream tasks. 


In this paper, we empirically study this potential issue of fairness with contrastive SSL approaches, and propose a new method, Conditional Contrastive Learning (CCL), to reduce the effect from a sensitive attribute during the contrastive pre-training. We focus on scenarios where the sensitive attribute is known and our goal is to mitigate its effect. The proposed CCL approach first defines the sensitive attribute (e.g., gender) as a conditional variable and samples the positive and negative pairs from distributions conditioning on the sensitive attribute. Empirically, this can be efficiently implemented by sampling from the same sensitive attribute, i.e., from the same gender. This simple but effective approach makes it harder for the model to leverage information from the sensitive attribute to distinguish positive pairs from negative pairs. We then proved that the proposed CCL maximizes a lower bound of \textit{conditional} mutual information between the learned representations, which explicitly excludes information from the conditional variable.



We evaluate our approach on five fairness datasets: Adult \citep{Dua:2019}, German \citep{Dua:2019}, COMPAS \citep{angwin2016machine}, Crime \citep{Dua:2019}, and Law School \citep{wightman1998lsac}, and two real-world facial datasets, CelebA \citep{liu2015faceattributes} and UTK-Face \citep{zhang2017age}. We study the fairness of contrastive SSL with multiple fairness metrics including demographic parity, equalized odds, and equality of opportunity, with the goal of maintaining strong downstream task performances compared to unsupervised baselines.

\section{Related Work}
\paragraph{Contrastive self-supervised learning.} Contrastive SSL has become successful in learning representations without labels \citep{radford2021learning, chen2020big, chen2020simple, riviere2020unsupervised}. The aim is to learn an embedding space that pulls together positive pairs and pushes away negative pairs \citep{chen2021empirical}. The representation can then be used for different downstream tasks, such as visual transfer learning \citep{chen2021empirical}, video action recognition \citep{radford2021learning}, geolocalization \citep{radford2021learning}, and speech recognition \citep{baevski2020wav2vec}. Recent work relates the success of contrastive self-supervised learning to mutual information maximization \citep{oord2018representation, bachman2019learning, bachman2019learning, chi2020infoxlm, tsai2021multiview}. Other works \citep{oord2018representation, poole2019variational, tosh2021contrastive, tsai2021multiview} have shown that contrastive learning objectives can be seen as maximizing the lower bound of mutual information between the two augmented views of the same image. 


Recent work in contrastive learning considers an additional conditional variable in contrastive learning to improve representation quality, such as auxiliary attributes \citep{tsai2021integrating}, information about the downstream task \citep{tian2020makes}, downstream labels \citep{khosla2020supervised, kang2020contragan} or data embeddings \citep{tsai2022conditional, wu2020conditional}. With the additional conditional variables, these works extend contrastive self-supervised learning to a weakly supervised \citep{tsai2021integrating}, semi-supervised \citep{tian2020makes}, or supervised setup \citep{khosla2020supervised}. This work, on the other hand, primarily aims to improve the fairness (instead of representation quality) in a self-supervised setup and proposes to use the given sensitive attribute from the dataset as the conditional variable.

Some work has discussed conditional distributions or conditional mutual information for contrastive learning. \citet{sordoni2021decomposed} proposes to capture more information between views in contrastive learning by decomposing mutual information into a sum of conditional mutual information, which conditions on subviews of the data (e.g., views obtained by occluding pixels of the original image).  Another work \citep{tsai2022conditional} approximates similarity scores from conditional distributions in contrastive learning using the conditional kernel mean embedding \citep{song2013kernel}. 
In comparison, this work also considers conditional mutual information and conditional distributions, but aims to remove (instead of capture) effect from an additional variable, and explicitly samples from conditional distributions (rather than using variational or kernel forms) to estimate conditional mutual information. 

\paragraph{Fair representation learning} To safely deploy self-supervised models in real-world scenarios such as healthcare, legal systems, and social science, it is also necessary to recognize the role they play in shaping social biases and stereotypes. Previous work has revealed that large-scale models trained with self-supervised learning can fail with respect to certain fairness criteria, such as models generating toxic speech~\cite{gehman2020realtoxicityprompts}, languages denigrating to particular social groups~\cite{liang2021towards,sheng2019woman}, among other concerns~\cite{blodgett2020language,hendrycks2020aligning,liang2020fair,sap2020social}. There has been work discussing fairness in self-supervised learning via reconstruction \citep{chakraborty2020fairmixrep}, but this work is the first to discuss the fairness in contrastive self-supervised learning for visual or tabular fairness (e.g., Adult \citep{Dua:2019}) datasets.

Recent work associates contrastive learning with fair representation learning. \citet{hong2021unbiased} improves fairness in a supervised setup by regularizing a contrastive term sampling positive pairs from the same downstream labels. \citep{shen2021contrastive} also considers a supervised setup with two contrastive objectives, where the positive pairs are from the same downstream label and the same sensitive attribute respectively, and the negative pairs are from different downstream labels or different sensitive attributes respectively. This work considers a self-supervised setup (instead of a supervised one), and samples both the positive and the negative pairs from the same sensitive attribute (i.e., the same gender). Recent work also proposes to maximize a lower bound of conditional mutual information (CMI) conditioning on the sensitive attribute to improve fairness, because CMI excludes information from the conditional variable \citep{mackay2003information}. \citet{song2019learning} considers an unsupervised setup and uses variational and adversarial objectives to estimate the CMI between the data and the representation. \citet{gupta2021controllable} considers a supervised setup and estimates the CMI between the downstream labels and the representations. In contrast, this work considers a self-supervised setup and estimates the CMI between representation of two data views using a contrastive objective.


\section{Method}

In this section, we introduce our proposed Conditional Contrastive Learning (CCL) approach to improve fairness in contrastive self-supervised learning, by reducing the impact of the given sensitive attribute in the representation. We first discuss the technical background, specifically focusing on conventional setup of contrastive self-supervised learning and the corresponding InfoNCE objective~\citep{oord2018representation}. We then present the CCL approach and motivate with theoretical results on why CCL can reduce the impact from the sensitive attribute.  

\subsection{Contrastive Self-Supervised Learning}

Conventional contrastive self-supervised learning~\citep{bachman2019learning,he2020momentum,chen2020simple} aims to learn an embedding space that pulls together representations of similar samples (positive pairs) and push away representations of dissimilar samples (negative pairs). For example, positive pairs could be two views of the same image by stochastic data augmentation \citep{chen2020simple}, while negative pairs could be two views of different images. Below, we first introduce notations of contrastive self-supervised learning, and then briefly summarize an information theory interpretation of conventional contrastive SSL \citep{oord2018representation, wu2020mutual, tsai2021multiview}, which help to both define and motivate our proposed CCL approach.

We use uppercase letters (e.g., $X$) to denote random variables and lowercase letters (e.g., $x$) to denote outcomes from the random variables. Specifically, we use $x$ and $y$ to denote the learned representations after encoding two data views $v_1$ and $v_2$ created by stochastic augmentation on images, where $v_1$ and $v_2$ could be a positive (from the same image) or a negative pair (from different images): 
\begin{equation}
\small{
    x= \text{encoder}\left(v_1\right), y=\text{encoder}\left(v_2\right),
}
\end{equation}
and $X$ and $Y$ are the corresponding random variables of representations $x$ and $y$. We use $P_X$ as the distribution of $X$ and $D_{\rm KL}\,\left(\cdot \,\|\,\cdot \right)$ as the Kullback–Leibler divergence between distributions.


Recent work \citep{oord2018representation, bachman2019learning, wu2020mutual, arora2019theoretical, tsai2021multiview} has shown that the success of contrastive SSL is related to maximizing a lower bound of mutual information shared between the representations of positive pairs. Specifically, the positive pair is sampled from the joint distribution, while the negative pairs are sampled from the product of marginal distributions. Mutual information is the KL divergence between the joint distribution and the product of marginal distributions, and optimizing a contrastive loss like \citet{oord2018representation} maximizes a lower bound of mutual information. Formally, the InfoNCE~\citep{oord2018representation} objective is to maximize $\text{MI}(X; Y)$ as follows:
\begin{equation}
\footnotesize
    \text{InfoNCE} :=\underset{f}{\rm sup}\,\,\mathbb{E}_{(x_i, y_i)\sim {P_{X,Y}}}\left[\,\frac{1}{n}\sum_{i=1}^n {\rm log}\,\frac{e^{f(x_i, y_i)}}{\frac{1}{n}\sum_{j=1}^n e^{f(x_i, y_j)}}\right] \leq D_{\rm KL}\,\left(P_{X,Y} \,\|\,P_{X}P_Y \right) = \text{MI}(X; Y),
\label{eq:infonce}
\end{equation}
where the positive pairs $\{(x_i, y_i)\}_{i=1}^n$ are  drawn from the joint distribution: $(x_i,y_i)\sim P_{X,Y}$, and the negative pairs $\{(x_i,y_{j \neq i})\}$ are drawn from the product of marginal distributions: $(x_i,y_{j \neq i}) \sim P_{X}P_{Y}$. $\text{MI}(X;Y)$ is the mutual information between $X$ and $Y$. $f(x,y)$ is any similarity scoring function that considers the input $(x,y)$ and output a similarity score. A common choice of $f(x,y)$ is the cosine similarity $f(x,y) = \cos{ \left( g(x), g(y) \right)} / \tau$, with $\tau$ being the temperature hyper-parameter and $g(\cdot)$ being a small neural network~\citep{he2020momentum,chen2020simple}. At a high-level, InfoNCE performs contrastive learning by maximizing the similarity for the positive pairs and minimizing the similarity for the negative pairs. As shown in \eqref{eq:infonce}, InfoNCE is a lower bound of $\text{MI}(X;Y)$, and several works \citep{arora2019theoretical, tsai2021multiview} have shown that maximizing lower bounds of $\text{MI}(X;Y)$ leads to better representations for downstream tasks. 

\subsection{Conditional Contrastive Learning}


Our proposed method, Conditional Contrastive Learning (CCL), differs from conventional contrastive SSL, by taking positive pairs and negative pairs from the distributions conditioning on a sensitive attribute referred as $Z$. Our CCL approach reduces information from the sensitive attribute $Z$ by taking $Z$ as the conditional variable between $X$ and $Y$. We focus on scenarios where $Z$ is readily available in the dataset (e.g., gender, race or age), following the practice of previous work on fairness \citep{madras2018learning, song2019learning}. Now we present the proposed \textbf{Conditional Contrastive Learning} objective: 

\begin{equation}
\small
 \text{CCL}: = \underset{f}{\rm sup}\,\,\mathbb{E}_{z\sim P_Z}\left[\mathbb{E}_{(x_i, y_i)\sim {P_{X,Y|z}}}\left[\,\frac{1}{n}\sum_{i=1}^n {\rm log}\,\frac{e^{f(x_i, y_i)}}{\frac{1}{n}\sum_{j=1}^n e^{f(x_i, y_j)}}\right]\right]
\label{eq:ccl}
\end{equation}
\vspace{1mm}

where the positive pairs $\{(x_i,y_i)\}_{i=1}^n$ represent samples drawn from the \textit{conditional} joint distribution: $(x_i,y_i) \sim P_{X,Y|z}$, while the negative pairs $\{(x_i,y_{j \neq i})\}$ represent samples drawn from the product of \textit{conditional} marginal distributions: $(x_i,y_{j \neq i}) \sim P_{X|z}P_{Y|z}$. The score function is $f(x,y) = \cos{ \left( g(x), g(y) \right)} / \tau$, same as \eqref{eq:infonce}. The difference between CCL and InfoNCE can be phrased as follows: CCL first samples $z \sim Z$, and then samples positive and negative pairs from $P_{X,Y|z}$ and $P_{X|z}P_{Y|z}$, respectively. InfoNCE, on the contrary, directly samples from $P_{X,Y}$ and $P_{X}P_{Y}$. Sampling from conditional distributions in CCL means that all positive and negative pairs share the same outcome $z$ of the sensitive attribute $Z$. Empirically, we could implement this by first sampling from the sensitive attribute (e.g., the gender), and then sampling positive and negative pairs from the same outcome of the sensitive attribute (i.e., the same gender). While we focus on InfoNCE for this paper, we should note that CCL can also be applied to other contrastive objective functions, especially divergence-based, such as Donsker-Varadhan \citep{donsker1975asymptotic}, Jensen-Shannon \citep{belghazi2018mutual}, or Wasserstein \citep{ozair2019wasserstein}, by applying the aforementioned sampling procedure to get the positive and negative pairs and plugging them in the contrastive objectives.



\subsection{Theoretical Motivation}
\label{method:theory}

In this section, we provide the theoretical motivation for our CCL method. In particular, we are interested in understanding why our method can reduce the information related to the sensitive attribute $Z$. Due to space limit, we defer detailed proofs of Equations \eqref{eq:cmi} and \eqref{eq:ccl_ineq} to Appendix Section A. Recall that InfoNCE is maximizing a lower bound of $D_{\rm KL}\,\left(P_{X,Y} \,\|\, P_{X}P_Y\right)$ as shown in \eqref{eq:infonce}. Similarly, our CCL method aims to maximize the divergence between $P_{XY|z}$ and $P_{X|z} P_{Y|z}$ for all $z\sim P_Z$, leading to a connection with conditional mutual information ${\rm MI}(X;Y|Z)$. First, we define conditional mutual information (CMI):
\begin{equation}
\small
\text{CMI}(X;Y|Z):= \mathbb {E} _{z\sim Z}\left[D_{\mathrm {KL} }\left(P_{X,Y|Z=z}\|P_{X|Z=z} P_{Y|Z=z}\right)\right] = \int _{\mathcal {Z}}D_{\mathrm {KL} }\left(P_{X,Y|Z}\|P_{X|Z} P_{Y|Z}\right)\,{\rm d}P_{Z}, \\
\label{eq:cmi}
\end{equation}
which measures the expected mutual information of $X$ and $Y$ given $Z$. Intuitively, $\text{CMI}(X;Y|Z)$ measures the averaged shared information by $X$ and $Y$ but exclude the effect from $Z$ \citep{mackay2003information}. This is because conditioning $Z=z$ means taking $Z=z$ as known and, therefore, ignoring the effect of $Z$ \citep{novovivcova2007conditional}. By ignoring the effect of $Z$, $\text{CMI}(X;Y|Z)$ explicitly excludes the information from $Z$ when measuring the shared information between $X$ and $Y$.

Next, we show our main theoretical result, that the proposed CCL objective is a lower bound of the conditional mutual information $\text{CMI}(X;Y|Z)$: 
\begin{equation}
\small
\text{CCL} \leq D_{\rm KL}\,\left(P_{X,Y} \,\|\,\mathbb{E}_{P_Z}\left[P_{X|Z}P_{Y|Z}\right] \right) = {\rm \text{Weak-CMI}}\,(X;Y|Z) \leq \text{CMI}(X;Y|Z),
\label{eq:ccl_ineq}
\end{equation}
where $\text{Weak}$-$\text{CMI}\,(X;Y|Z)$ is the KL-divergence between $P_{X,Y}$ and $\mathbb{E}_{P_Z}\left[P_{X|Z}P_{Y|Z}\right]$. This notion has been used to achieve the so-called weak-conditional independence~\citep{daudin1980partial,fukumizu2004dimensionality,fukumizu2007kernel}. We have the weak conditional independence between $X$ and $Y$ given $Z$ when $\text{Weak}$-$\text{CMI}\,(X;Y|Z) = 0$. First, $\text{Weak}$-$\text{CMI}\,(X;Y|Z) = 0$ is a necessary but not sufficient condition for $\text{CMI}(X;Y|Z) = 0$, suggesting that conditional independence implies weak conditional independence. For example, if $X$, $Y$, and $Z$ are pairwise independent but jointly dependent, $\text{Weak}$-$\text{CMI}\,(X;Y|Z) = 0$ but $\text{CMI}(X;Y|Z)$ may not be zero. Although weak conditional independence does not fully characterize conditional independence, it has been shown to be widely useful in practice. For instance, testing weak conditional independence can be simpler and more powerful than the original conditional independence test \citep{zhang2017feature}. Our approach benefits from the notion of weak conditional independence in similar ways. Also, we prove that $\text{Weak}$-$\text{CMI}\,(X;Y|Z)$ is a lower bound of $\text{CMI}(X;Y|Z)$ and can be seen as a more ``conservative'' measurement of $\text{CMI}(X;Y|Z)$, capturing only part of information in $\text{CMI}(X;Y|Z)$.

\begin{wrapfigure}{r}{0.4\textwidth}
\centering
  \includegraphics[width=1.3\linewidth]{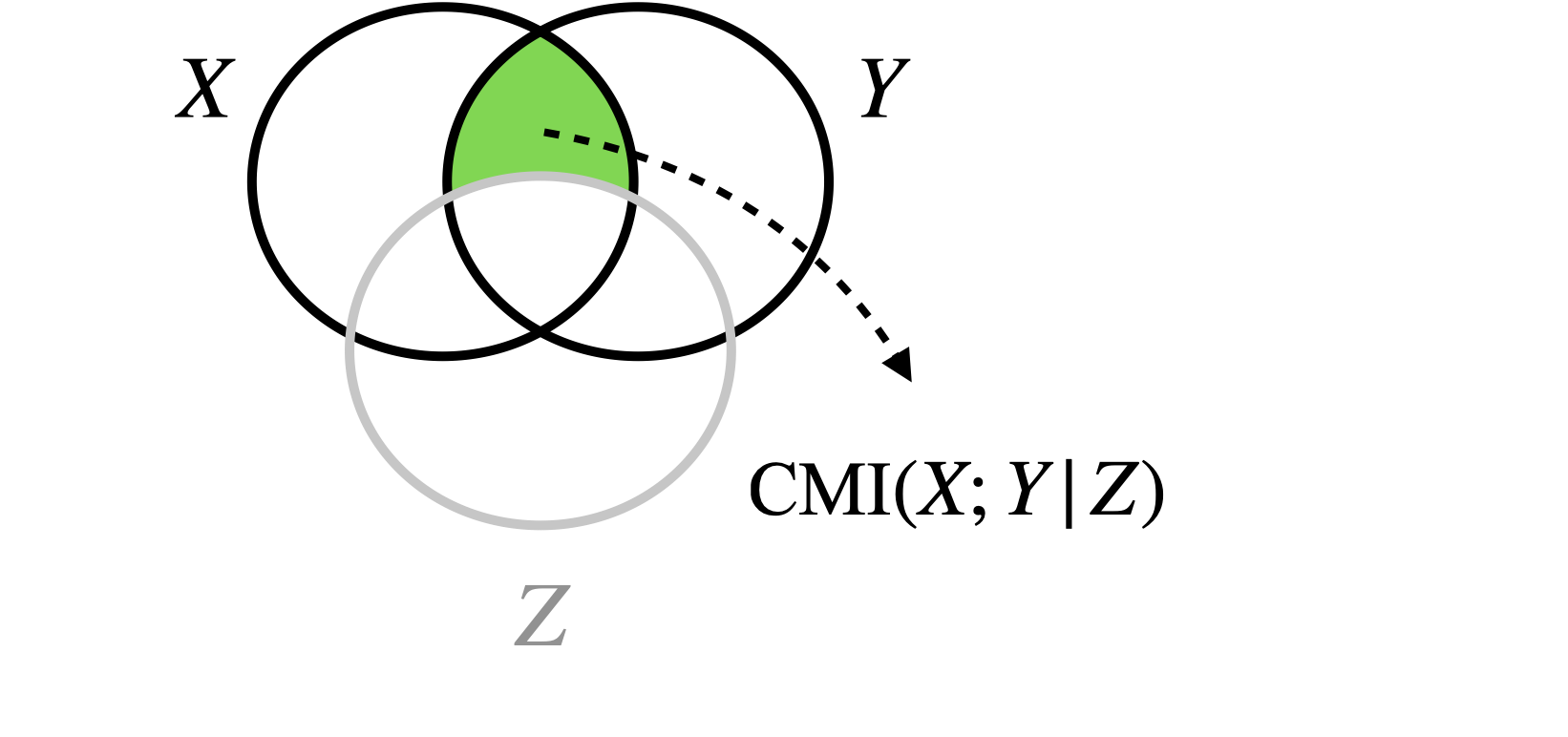}
\vspace{-4mm}
\caption{Venn diagram of $\text{CMI}(X;Y|Z)$ (the green section).}
\label{fig:venn}
\end{wrapfigure}
\vspace{-2mm}

\paragraph{Why CCL may work?} Two main observations can be deduced from the theoretical result. The first observation relates to \textit{fairness}, while the second observation relates to \textit{representation quality}. For the first observation concerning fairness, we draw a Venn diagram of $\text{CMI}\,(X;Y|Z)$ to illustrate why the impact of $Z$ is reduced. As shown in Figure \ref{fig:venn}, $\text{CMI}\,(X;Y|Z)$ explicitly excludes information from $Z$ \citep{mackay2003information}. Since CCL is the lower bound of $\text{CMI}(X;Y|Z)$, the impact of $Z$ will be reduced as we optimize CCL. For the second observation about representation quality, maximizing CCL results in maximizing a lower bound of $\text{CMI}(X;Y|Z)$ between the representation $X$ of data view $V_1$ and representation $Y$ of data view $V_2$ given $Z$. Previous work such \citep{hjelm2018learning, oord2018representation, tsai2021multiview} has shown that maximizing the information shared between $X$ and $Y$ can produce a good embedding space that has high \textit{representation quality} for downstream tasks.





\section{Experiments}

We evaluate the proposed Conditional Contrastive Learning on several tasks, summarized in Table \ref{tab:dataset}. We experiment with five fairness datasets: Adult \citep{Dua:2019},  Compas \citep{angwin2016machine}, Crime \citep{Dua:2019}, German \citep{Dua:2019}, and Law School \citep{wightman1998lsac}, and two facial datasets: CelebA \citep{liu2015deep} and UTKFace \citep{zhang2017age}. The sensitive attributes for each experiment are also summarized in Table \ref{tab:dataset}. The corresponding sensitive attribute for each dataset is used as the conditioning variable for the proposed CCL. We evaluate on prediction accuracy (all tasks being binary predictions) and three fairness metrics (in the form of distance): Demographic Parity ($\Delta_{DP}$), Equalized Odds ($\Delta_{EO}$), and Equality of Opportunity ( $\Delta_{{EO}_{PP}}$). We include all implementation details, including hyperparameters, datasets and source code in Appendix Section B. 

\begin{wraptable}{l}{0.7\linewidth}
\caption{Details of datasets, the chosen sensitive attributes, and the corresponding prediction tasks. There are three prediction task for CelebA: attractiveness, weary hair, and smiling. }

\centering
\resizebox{\linewidth}{!}{
	\begin{tabular}{lllll}
		\toprule 
		\multirow{2}{*}{Datasets} &\multirow{2}{*}{Type} & Number &Sensitive& Prediction  \\
		& &  of Samples & Attribute & Task(s)\\
		\midrule
		Adult \citep{Dua:2019} & Tabular & $48,842$ & Gender & Income level \\
		Compas \citep{angwin2016machine}&Tabular & $5,278$  & Race & Recidivism \\
		Crime \citep{Dua:2019}& Tabular & $1,994$ & Race & Crime level \\
		German \citep{Dua:2019}& Tabular & $1,000$  & Age & Credit approval \\
		Law School \citep{wightman1998lsac}& Tabular & $36,022$ & Race & Exam result \\
		UTKFace \citep{zhang2017age} & Vision & $23,708$ & Race & Age \\
		CelebA \citep{liu2015deep}& Vision & $202,599$& Gender& Multiple \\
		\bottomrule
	\end{tabular}
}
	\label{tab:dataset}
\end{wraptable}
\subsection{Fairness Criteria}

We use three types of fairness metrics: the demographic parity (DP \citep{feldman2015certifying}) distance $\Delta_{DP}$ ~\citep{madras2018learning}, equalized odds (EO \citep{hardt2016equality}) distance $\Delta_{EO}$ \citep{song2019learning}, and the equality of opportunity ($EO_{PP}$ \citep{hardt2016equality}) distance $\Delta_{{EO}_{PP}}$ \citep{song2019learning}.  Given the data $X$, the sensitive attribute $Z$ indicating group information, the ground truth downstream task label $l$, and the label prediction from the model $\hat{l}$, the $\Delta_{DP}$  calculates the expected difference (in absolute value) in model predictions between two groups: $\Delta_{DP} = |\mathbb{P}\{\hat{l}=1 | Z = 0\} - \mathbb{P}\{\hat{l}=1 | Z = 1\}|$. The second metric, $\Delta_{EO}$, calculates the sum of the expected difference (in absolute value) of the True Positive Rate and the False Positive Rate of the model predictions between two groups: $\Delta_{EO} = |\mathbb{P}\{\hat{l}=1 | Z = 0, l=1 \} - \mathbb{P}\{\hat{l}=1 | Z = 1, l=1\}| + |\mathbb{P}\{\hat{l}=1 | Z = 0, l= 0 \} - \mathbb{P}\{\hat{l}=1 | Z = 1, l=0\}$. As a relaxation of $\Delta_{EO}$, $\Delta_{{EO}_{PP}}$ calculates the expected difference (in absolute value) of only the True Positive Rate of the model predictions between two groups: $\Delta_{EO_{PP}} = |\mathbb{P}\{\hat{l}=1 | Z = 0, l=1 \} - \mathbb{P}\{\hat{l}=1 | Z = 1, l=1\}|$. $\Delta_{DP}$, $\Delta_{EO}$,  $\Delta_{{EO}_{PP}}$ range from 0 to 1, and a smaller distance is desirable. $\Delta_{DP} = 0$ corresponds to the statistical independence of the sensitive attribute $Z$ and the prediction $\hat{l}$, and $\Delta_{EO} = 0$ corresponds to the conditional independence of $Z$ and $\hat{l}$ given the true label $l$. Intuitively, for example, $\Delta_{DP} = 0$ suggests that members of different groups (e.g., female and male) have the same chance of receiving a favorable prediction ($l=1$). 

%

\subsection{Experimental Methodology}
We follow the setup from the contrastive SSL learning literature \citep{chen2020simple, he2020momentum}, which contains two stages: contrastive pre-training and supervised fine-tuning. We use the SimCLR framework \citep{chen2020simple}. In contrastive pre-training, we train an encoder without any labels. In the supervised fine-tuning stage, we freeze the encoder and fune-tine an additional small network with the downstream labels. We then evaluate the fine-tuned representations on the test splits of the corresponding datasets. For fairness datasets, we use a three-layer neural network with hidden dimension $100$ as the encoder and a linear layer as the fine-tuning network. For vision datasets, we use a ResNet-50~\cite{he2016deep} as the encoder and a two-layer network as the fine-tuning network. There are two types of baselines: unsupervised and SSL baselines. Unsupervised baselines include models dedicated for improving fairness in unsupervised representations. The SSL baseline includes implementations of the InfoNCE loss on SimCLR. We did not include supervised fair representation models, as they often require labels and sensitive attributes to be available at the same time, which is not our case.

\subsection{Fairness Dataset Experiments}
\paragraph{Implementation Details.} The self-supervised baseline, SimCLR \citep{oord2018representation}, and the unsupervised baseline, LCIFR \citep{ruoss2020learning} are re-implemented based on \citet{ruoss2020learning}. To stochastically augment tabular features (e.g., age, education, occupation, etc in Adult \citep{Dua:2019} dataset) and create data views similar to \citet{chen2020simple}, we first standardize each tabular feature, and then use noise vectors from an isotropic Gaussian to perturb the features. Each dataset uses one separate Gaussian, and $\sigma$ of the Gaussian is treated as a hyper-parameter for different datasets. Then we feed the augmented views to the encoder, and then use the output of the encoder to estimate our proposed CCL.
\begin{table}[H]
\centering
\caption{Accuracies and fairness results on five fairness datasets. Details of these baselines are in Appendix Section B. Best results are in bold. CCL has better downstream accuracy than existing unsupervised and self-supervised baselines in four datasets, and exhibits better fairness measurements in 11 out of 18 results.}

\begin{adjustbox}{center,width=0.95\linewidth}
\begin{tabular}{lllcccc}
\toprule

&&Model &Accuracy (\%) $(\uparrow)$ & $\Delta_{DP}\,(\downarrow)$ & $\Delta_{EO}\,(\downarrow)$ & $\Delta_{{EO}_{PP}}\,(\downarrow)$      \\ 
\midrule

\multicolumn{2}{c}{\parbox[t]{2mm}{\multirow{9}{*}{\rotatebox[origin=c]{90}{ADULT}}}} & \bf Unsupervised\\
&&--  LAFTR \citep{madras2018learning} & 84.0 & 0.163 & \bf 0.030 & \bf 0.026 \\
&&-- \citet{ragonesi2021learning} & 85.0 & - & \bf 0.030 & -  \\
&&-- DTM \citep{lee2020maximal} & 71.6 & -& 0.050 & -\\
&&-- FNF \citep{balunovic2021fair} & 80.0 & \bf 0.110 & - & - \\
&&\bf Self-Supervised & \\
&&-- SimCLR \citep{chen2020simple} & 83.1 & 0.210 & 0.410 & 0.320\\
&&-- FairMixRep \citep{chakraborty2020fairmixrep}  & 85.0 & 0.172 & - & - \\
&&-- \bf CCL (Ours) & \bf 85.4 & \bf 0.110 & 0.070  & 0.090 \\

\midrule
\multicolumn{2}{c}{\parbox[t]{2mm}{\multirow{6}{*}{\rotatebox[origin=c]{90}{COMPAS}}}} &\bf Unsupervised \\
&&-- DTM \citep{lee2020maximal} & 66.0 & - & 0.200 & - \\
&&-- FNF \citep{balunovic2021fair} & 65.0 & 0.240 & - & - \\
&&\bf Self-Supervised & \\
&&-- SimCLR \citep{chen2020simple}  & \bf 71.2 & 0.103 & 0.227 & 0.134 \\
&&-- \bf CCL (Ours) & 71.0& \bf 0.080 & \bf 0.132  & \bf 0.081 \\


\midrule
\multicolumn{2}{c}{\parbox[t]{2mm}{\multirow{6}{*}{\rotatebox[origin=c]{90}{CRIME}}}}&\bf Unsupervised&\\

&&-- LCIFR \citep{ruoss2020learning} & \bf 84.4 & 0.443 & 0.314 & 0.212 \\
&&-- FNF \citep{balunovic2021fair} & 82.5 & 0.540 & - & - \\
&&\bf Self-Supervised\\
&&-- SimCLR \citep{chen2020simple} & 82.1 & 0.502 & 0.530 & 0.383 \\
&&-- \bf CCL (Ours) & 82.6 & \bf 0.211 & \bf 0.224  & \bf 0.183 \\

\midrule
\multicolumn{2}{c}{\parbox[t]{2mm}{\multirow{7}{*}{\rotatebox[origin=c]{90}{GERMAN}}}}& \bf Unsupervised\\
&&-- LCIFR \citep{ruoss2020learning} & 73.1 & 0.102 & 0.080 & 0.063 \\

&&-- \citet{ragonesi2021learning} & 74.0 & - & \bf 0.060 & -  \\
&&\bf Self-Supervised\\
&&-- FairMixRep \citep{chakraborty2020fairmixrep}  & 71.8 & 0.089 & - & - \\

&&-- SimCLR \citep{chen2020simple}   & 72.5 &0.250 & 0.382 & 0.195 \\
&&-- \bf CCL (Ours) & \bf 74.3 & \bf 0.083 & 0.128  & \bf 0.062 \\

\midrule
\parbox[t]{0.1mm}{\multirow{6}{*}{\rotatebox[origin=c]{90}{LAW}}} & \parbox[t]{2mm}{\multirow{6}{*}{\rotatebox[origin=c]{90}{SCHOOL}}} &
\bf Unsupervised\\ 
&&-- LCIFR \citep{ruoss2020learning} & 84.4 & 0.110 & 0.180 & 0.070 \\
&&-- FNF \citep{balunovic2021fair} & 84.6 & \bf 0.050 & - & - \\
&&\bf Self-Supervised\\
&&-- SimCLR \citep{chen2020simple}   & 83.6 & 0.086 & 0.212 & 0.110 \\
&&-- \bf CCL (Ours) & \bf 84.8 & 0.051 & \bf 0.153  & \bf 0.056\\


\bottomrule
\end{tabular}
\end{adjustbox}
\label{tab:fair}

\end{table}


\paragraph{Results.} Table \ref{tab:fair} shows the results on accuracy and fairness metrics for both unsupervised and self-supervised methods. First, we observe that self-supervised SimCLR and CCL have strong downstream prediction results close to or better than the state-of-the-art baselines in Adult, Compas, German and the Law School datasets. Next, looking at the fairness measurements, we observe that the SimCLR baseline performs significantly worse than unsupervised baselines, sometimes two to three times higher ($\Delta_{DP}$ in Adult, $\Delta_{EO}$ in German, and $\Delta_{{EO}_{PP}}$ in Adult), confirming our earlier concern that contrastive self-supervised learning will produce highly unfair predictions without bias mitigation. The proposed CCL is much better than SimCLR and very competitive compared to other unsupervised baselines, in terms of fairness criteria: the average improvement over five datasets from SimCLR to CCL is $12.28\%$ on $\Delta_{DP}$, $21.08\%$ on $\Delta_{EO}$, and $13.43\%$ on $\Delta_{{EO}_{PP}}$.

\begin{wrapfigure}{r}{0.55\textwidth}
\centering
\begin{minipage}{.5\textwidth}
  \centering
  \includegraphics[width=\linewidth]{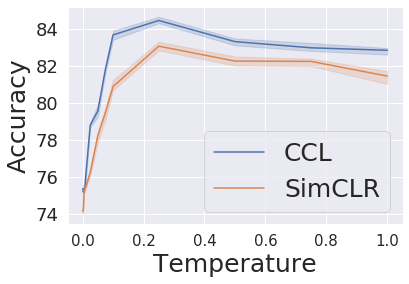}
\end{minipage}%
\begin{minipage}{.5\textwidth}
  \centering
  \includegraphics[width=\linewidth]{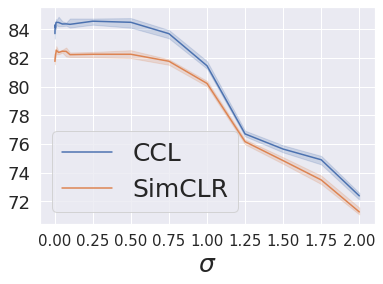}
\end{minipage}
\caption{Accuracy using different augmentation noise level $\sigma$ and temperature $\tau$. Left: Varying temperature $\tau$ on prediction accuracy, $\sigma=0.25$. Right: Varying augmentation noise $\sigma$ on prediction accuracy, $\tau=0.25$. A mid-range $\tau=0.25$ and a mild noise level $\sigma=0.25$ helps the most for learning strong representations.}
\vspace{-5mm}
\label{fig:ablation_fair}
\end{wrapfigure}

\paragraph{Effect of hyper-parameters on downstream performances.} We study two important hyper-parameters using the Adult dataset: the $\sigma$ of the Gaussian noise and the temperature $\tau$ in Equation \ref{eq:ccl}. The Gaussian noise controls the level of data augmentation, and $\tau$ smooths the distribution of the score output of the encoder. Both will influence the representation quality in contrastive learning. We use $\tau \in [0.001, 1]$; and $\sigma \in [0.001, 2]$. The results are shown in Figure \ref{fig:ablation_fair}. We observe that a mid-range $\tau=0.25$ achieves the best results. The prediction accuracy begins to increase drastically as $\tau$ goes from $0.001$, tops at $\tau=0.25$, and start decreasing slightly from $\tau=0.5$. Next, for the noise level, we observe that a small $\sigma$ ranging from $0.001$ to $0.75$ achieves similar results (around or above $84\%$), peaks at $\sigma=0.25$, and then degrades fast after $0.75$. This suggests that a mid-range temperature ($\tau=0.25$) to smooth the similarity score distribution \citep{hinton2015distilling}, and a mild noise augmentation ($\sigma=0.25$) to the tabular data help the most in representation learning. 

\paragraph{Effect of hyper-parameters on fairness.} We also study the effect of $\sigma$ and $\tau$ in terms of fairness criteria: $\Delta_{DP}$, $\Delta_{EO}$, and $\Delta_{{EO}_{PP}}$. Overall, a similar trend occurs for three criteria: a large noise ($\sigma > 0.75$) and a large temperature ($\tau > 0.5$) generates the worst representation in terms of fairness metrics ( $\Delta_{DP} > 0.2$, $\Delta_{EO} > 0.3$ and $\Delta_{{EO}_{PP}} > 0.3$). On the other hand, a large noise ($\sigma > 0.75$) and a medium-to-small temperature ($\tau < 0.5$) generates the best results on fairness, but in these cases the representation performs badly on downstream tasks. The right trade-off between representation power and fairness we found is $\tau \in [0.1, 0.5]$ and $\sigma \in [0.001, 0.25]$. The results suggest that a larger noise and a mid-range temperature may help remove bias information.

\subsection{Vision Dataset Experiments}

\begin{wrapfigure}{r}{0.45\textwidth}
\centering
\begin{minipage}{.5\textwidth}
  \centering
  \includegraphics[width=1.1\textwidth]{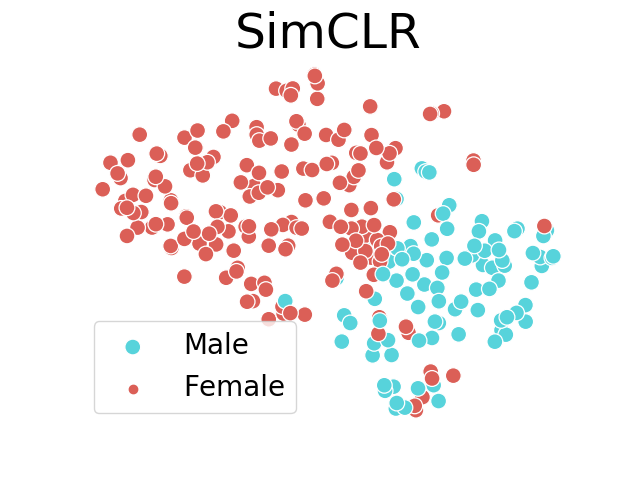}
  \label{fig:ablation_temp}
\end{minipage}%
\begin{minipage}{.5\textwidth}
  \centering
  \includegraphics[width=1.1\linewidth]{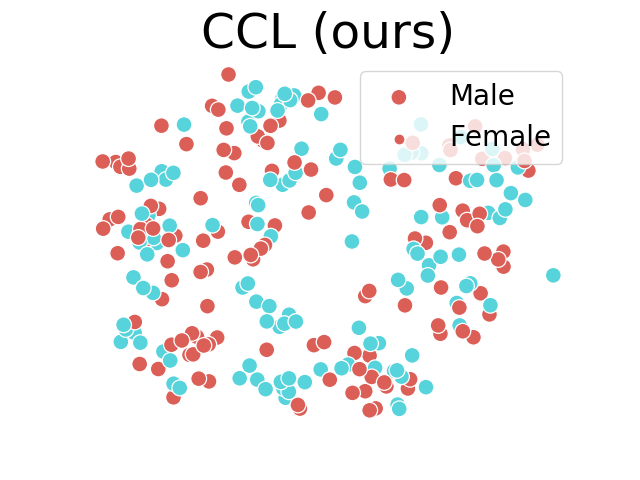}
  \label{fig:ablation_sigma}
\end{minipage}
\vspace{2mm}
\caption{\small{T-SNE embeddings of conventional contrastive SSL (SimCLR, left) vs. Conditional Contrastive Learning (ours, right). Gender groups are visually inseparable in the CCL (right) compared to conventional contrastive SSL (left), suggesting an embedding with less gender biases.}}
\vspace{-8mm}

\label{fig:ablation_vision}
\end{wrapfigure}

We implement the baseline SimCLR by following \citet{chen2020simple} and using the augmentations resize-and-crop, color jitter and horizontal flip. We evaluate on the prediction tasks specified in Table \ref{tab:dataset} (age prediction for UTKFace, and attractiveness / weary hair / smiling prediction for CelebA). The sensitive attribute for UTKFace is race, and gender for CelebA. The results are in Table \ref{tab:fair}. We observe that the proposed CCL both outperforms unsupervised baselines and the self-supervised SimCLR baseline on the prediction tasks. The CCL also achieves much better fairness criteria than the SimCLR baseline, producing much lower $\Delta_{DP}$, $\Delta_{EO}$,  and $\Delta_{{EO}_{PP}}$ in all four tasks across two datasets.

We also plot the embedding spaces of SimCLR and CCL using t-SNE \citep{van2008visualizing}. The visualization is in Figure \ref{fig:ablation_vision}. The embeddings of female and male group samples are clearly separated in SimCLR, making it easy for downstream fine-tuning models to pick up gender information and produce unfair predictions. On the other hand, for the embedding from CCL it is much hard to separate two groups, making it hard for models to leverage gender information from the representation.

\vspace{-1mm}
\begin{wraptable}{l}{0.6\linewidth}
\caption{\small Fairness criteria on contrastive SSL vs. supervised counterpart. Contrastive SSL has mugh higher level of fairness differences based on the three metrics.}
\vspace{-2mm}
\centering
\scalebox{0.75}{
\begin{tabular}{ccccccc}
\toprule
& Accuracy & \textbf{$\Delta_{DP}$ ($\downarrow$)} & \textbf{$\Delta_{EO}$ ($\downarrow$)} & \textbf{$\Delta_{{EO}_{PP}}$ ($\downarrow$)}
\\ \midrule
Supervised & 80.4 & 0.214 & 0.186 & 0.080 \\
Contrastive SSL & 80.1 & 0.355 & 0.541 & 0.310 \\
 \bottomrule
\end{tabular}
}
\label{tab:worse}
\end{wraptable}
\label{sec:behavior}

\textbf{Contrastive SSL performs worse on fairness than supervised methods.} To study whether contrastive SSL models perform better or worse on fairness criteria than a supervised counterpart, we train two ResNet-18 models~\citep{he2016deep}, one with contrastive pre-training then fine-tuning \citep{he2020momentum}, and one with supervised training. Both models have the same architecture and training hyperparameters. From Table \ref{tab:worse}, given similar performance, contrastive SSL has significantly larger fairness differences, suggesting that contrastive SSL can produce downstream predictions that perform much worse on fairness criteria than its supervised counterpart.
\begin{table}[H]
\centering
\caption{Accuracies and fairness results on two vision datasets with four prediction tasks. Best results are bold. CCL has better downstream accuracy in all four tasks, and exhibits better or close-to-the-best fairness measurements in 11 out of 12 results.}

\begin{adjustbox}{center,width=0.95\linewidth}
\begin{tabular}{lllcccc}
\toprule

&&Model &Accuracy (\%) $(\uparrow)$ & $\Delta_{DP}\,(\downarrow)$ & $\Delta_{EO}\,(\downarrow)$ & $\Delta_{{EO}_{PP}}\,(\downarrow)$      \\ 
\midrule

\parbox[t]{0.1mm}{\multirow{7}{*}{\rotatebox[origin=c]{90}{CELEBA}}} & \parbox[t]{2mm}{\multirow{7}{*}{\rotatebox[origin=c]{90}{ATTRACTIVE}}} &
\bf Unsupervised\\ 
&&-- MFD \citep{jung2021fair} & 80.2 & - & \bf 0.050 & - \\
&&-- \citet{balunovic2021fair} & 79.4 & - & 0.238 & - \\
&&-- \citet{morales2020sensitivenets} & 77.7 & - & 0.070 & - \\
&&\bf Self-Supervised\\
&&-- SimCLR \citep{chen2020simple}   & 81.7 & 0.277 & 0.212 & 0.110 \\
&&-- \bf CCL (Ours) & \bf 82.1 & \bf 0.202 & 0.101  & \bf 0.048\\
\midrule
\parbox[t]{0.1mm}{\multirow{6}{*}{\rotatebox[origin=c]{90}{CELEBA}}} & \parbox[t]{2mm}{\multirow{6}{*}{\rotatebox[origin=c]{90}{WAVY HAIR}}} &
\bf Unsupervised\\ 
&&-- FactorVAE \citep{kim2018disentangling} & 64.5 & - & 0.388 & 0.288 \\
&&-- FFVAE \citep{creager2019flexibly} & 61.0 & - & 0.211 & 0.154 \\
&&\bf Self-Supervised\\
&&-- SimCLR \citep{chen2020simple}   & \bf 67.7 & 0.403 & 0.355 & 0.210 \\
&&-- \bf CCL (Ours) & \bf 67.7 & \bf 0.202 & \bf 0.189  & \bf 0.102\\

\midrule
\parbox[t]{0.1mm}{\multirow{5}{*}{\rotatebox[origin=c]{90}{CELEBA}}} & \parbox[t]{2mm}{\multirow{5}{*}{\rotatebox[origin=c]{90}{SMILE}}} &
\bf Unsupervised\\ 
&&-- \citet{morales2020sensitivenets} & 88.4 & - & \bf 0.060 & - \\

&&\bf Self-Supervised\\
&&-- SimCLR \citep{chen2020simple}   & 89.3 & 0.102 & 0.142 & 0.078 \\
&&-- \bf CCL (Ours) & \bf 89.7 & \bf 0.086 & \bf 0.060  & \bf 0.053\\

\midrule
\parbox[t]{0.1mm}{\multirow{6}{*}{\rotatebox[origin=c]{90}{UTKFACE}}} & \parbox[t]{2mm}{\multirow{6}{*}{\rotatebox[origin=c]{90}{GENDER}}} &
\bf Unsupervised\\ 
&&-- AD \citep{zhang2018mitigating} & 74.7 & - & 0.204 & - \\
&&-- MFD \citep{jung2021fair} & 74.7 & - & 0.178 & - \\
&&\bf Self-Supervised\\
&&-- SimCLR \citep{chen2020simple}   & 78.0 & 0.335 & 0.421 & 0.287 \\
&&-- \bf CCL (Ours) & \bf 78.5 & \bf 0.191 & \bf 0.156  & \bf 0.089\\

\bottomrule
\end{tabular}
\end{adjustbox}
\label{tab:vision}

\end{table}

\vspace{-1mm}
\section{Discussion, Limitations and Social Impact}
\vspace{-1mm}

We introduce Conditional Contrastive Learning (CCL) which samples positive and negative pairs from distributions conditioning on the sensitive attribute to remove its effect, and thus improving fairness in self-supervised learning. By conditioning on the sensitive attribute, the positive and negative pairs come from the same subgroup, making it harder for the model to leverage gender-related information. We prove that CCL is a lower bound of conditional mutual information, and optimizing it leads to learning strong representations for downstream tasks while reducing the information from the sensitive attribute. Empirically, we show that CCL significantly improves the fairness of conventional contrastive models, while achieving SOTA downstream performances compared to both contrastive SSL and other unsupervised baselines.

One important future work direction, and a limitation of this work, is to study the scenario the sensitive information is unknown or partially known. It could be addressed by using other auxiliary attributes (e.g., image annotations or captions) in the datasets that are highly relevant to the sensitive attributes, or first train a separate model to capture bias features and then train the main model by learning features orthogonal to the bias feature. Another important problem is to remove the effect of multiple sensitive attributes simultaneously, which may be addressed by using a joint distribution of multiple sensitive attributes. If there are too many sensitive attributes, we can perform a dimensional reduction. 
For the social impact, CCL may bring a positive impact by removing gender, race, or identity information from representations. The potential negative impact is that this method could be intentionally used to remove information that should be available and included in the representation, for example, gender information in a model for medical diagnosis.

\bibliographystyle{abbrvnat}
\bibliography{neurips_2022}

\clearpage

\appendix

\section{Theoretical Analysis}
This section provides the theoretical analysis of Equations (4) and (5) in the main text. The full set of assumptions of all theoretical results and complete proofs of all theoretical results are presented below.

\subsection{Useful lemmas}

We first present the following lemmas, which will be later used in the proof:

\begin{lemma}[\citet{nguyen2010estimating} with two variables] Let $\mathcal{X}$ and $\mathcal{Y}$ be the sample spaces for $X$ and $Y$, $f$ be any function: $(\mathcal{X} \times \mathcal{Y}) \rightarrow \mathbb{R}$, and $\mathcal{P}$ and $\mathcal{Q}$ be the probability measures on $\mathcal{X} \times \mathcal{Y}$. Then,
$$
D_{\rm KL} \left( \mathcal{P} \,\|\, \mathcal{Q} \right) = \underset{f}{\rm sup} \,\mathbb{E}_{(x, y)\sim\mathcal{P}} [f(x,y)] - \mathbb{E}_{(x, y)\sim\mathcal{Q}} [e^{f(x,y)}] + 1.
$$
\begin{proof}
The second-order functional derivative of the objective is $-e^{f(x,y)}\cdot d\mathcal{Q}$, which is always negative. The negative second-order functional derivative implies the objective has a supreme value. 
Then, take the first-order functional derivative and set it to zero:
\begin{equation*}
d \mathcal{P} - e^{f(x,y)}\cdot d \mathcal{Q} = 0.
\end{equation*}
We then get the optimal $f^*(x,y) = {\rm log}\,\frac{d\mathcal{P}}{d\mathcal{Q}}$. Plug in $f^*(x,y)$ into the objective, we obtain
\begin{equation*}
\mathbb{E}_{\mathcal{P}} [f^*(x,y)] - \mathbb{E}_{\mathcal{Q}} [e^{f^*(x,y)}] + 1 = \mathbb{E}_{\mathcal{P}} [{\rm log}\,\frac{d\mathcal{P}}{d\mathcal{Q}}] = D_{\rm KL} \left( \mathcal{P} \,\|\, \mathcal{Q} \right).
\end{equation*}
\end{proof}
\label{lemm:kl}
\end{lemma}

\begin{lemma}[\citet{nguyen2010estimating} with three variables] Let $\mathcal{X}$, $\mathcal{Y}$, and $\mathcal{Z}$ be the sample spaces for $X$, $Y$, and $Y$, $f$ be any function: $(\mathcal{X} \times \mathcal{Y} \times \mathcal{Z}) \rightarrow \mathbb{R}$, and $\mathcal{P}$ and $\mathcal{Q}$ be the probability measures on $\mathcal{X} \times \mathcal{Y} \times \mathcal{Z}$ . Then,
$$
D_{\rm KL} \left( \mathcal{P} \,\|\, \mathcal{Q} \right) = \underset{f}{\rm sup} \,\mathbb{E}_{(x, y, z)\sim\mathcal{P}} [f(x,y, z)] - \mathbb{E}_{(x, y, z)\sim\mathcal{Q}} [e^{f(x,y, z)}] + 1.
$$
\begin{proof}
The second-order functional derivative of the objective is $-e^{f(x,y, z)}\cdot d\mathcal{Q}$, which is always negative. The negative second-order functional derivative implies the objective has a supreme value. 
Then, take the first-order functional derivative and set it to zero:
\begin{equation*}
d \mathcal{P} - e^{f(x,y, z)}\cdot d \mathcal{Q} = 0.
\end{equation*}
We then get the optimal $f^*(x,y, z) = {\rm log}\,\frac{d\mathcal{P}}{d\mathcal{Q}}$. Plug in $f^*(x,y, z)$ into the objective, we obtain
\begin{equation*}
\mathbb{E}_{\mathcal{P}} [f^*(x,y, z)] - \mathbb{E}_{\mathcal{Q}} [e^{f^*(x,y, z)}] + 1 = \mathbb{E}_{\mathcal{P}} [{\rm log}\,\frac{d\mathcal{P}}{d\mathcal{Q}}] = D_{\rm KL} \left( \mathcal{P} \,\|\, \mathcal{Q} \right).
\end{equation*}
\end{proof}
\label{lemm:kl_3}
\end{lemma}

\subsubsection{Immediate results following Lemma~\ref{lemm:kl}}

\begin{lemma}
\begin{equation*}
\begin{split}
    {\rm \text{Weak-CMI}}\,(X;Y|Z)  & = D_{\rm KL} \left( P_{X,Y} \,\|\, \mathbb{E}_{P_Z}\left[P_{X|Z}P_{Y|Z}\right] \right)
    \\
    & = \underset{f}{\rm sup} \,\mathbb{E}_{(x, y)\sim P_{X,Y}} [f(x,y)] - \mathbb{E}_{(x, y)\sim \mathbb{E}_{P_Z}\left[P_{X|Z}P_{Y|Z}\right]} [e^{f(x,y)}] + 1.
\end{split}
\end{equation*}
\begin{proof}
Let $\mathcal{P}$ be $P_{X,Y}$ and $\mathcal{Q}$ be $\mathbb{E}_{P_Z}\left[P_{X|Z}P_{Y|Z}\right]$ in Lemma~\ref{lemm:kl}.
\end{proof}
\label{lemm:wc_infonce}
\end{lemma}

\begin{lemma} 
$
\underset{f}{\rm sup}\,\,\mathbb{E}_{(x, y_1)\sim \mathcal{P}, (x, y_{2:n})\sim \mathcal{Q}^{\otimes (n-1)}}\left[ {\rm log}\,\frac{e^{f(x, y_1)}}{\frac{1}{n}\sum_{j=1}^n e^{f(x, y_j)}}\right] \leq D_{\rm KL} \left( \mathcal{P} \,\|\, \mathcal{Q} \right).
$
\begin{proof}
$\forall f$, we have
\begin{equation*}
\small
\begin{split}
    D_{\rm KL} \left( \mathcal{P} \,\|\, \mathcal{Q} \right) & = \mathbb{E}_{(x, y_{2:n})\sim \mathcal{Q}^{\otimes (n-1)}}\left[D_{\rm KL} \left( \mathcal{P} \,\|\, \mathcal{Q} \right) \right]
    \\
    & \geq  \,\mathbb{E}_{(x, y_{2:n})\sim \mathcal{Q}^{\otimes (n-1)}}\left[ \mathbb{E}_{(x, y_1)\sim \mathcal{P}} \left[ {\rm log}\,\frac{e^{f(x, y_1)}}{\frac{1}{n}\sum_{j=1}^n e^{f(x, y_j)}} \right] - \mathbb{E}_{(x, y_1)\sim \mathcal{Q}} \left[ \frac{e^{f(x, y_1)}}{\frac{1}{n}\sum_{j=1}^n e^{f(x, y_j)}} \right] + 1 \right] \\
    & = \mathbb{E}_{(x, y_{2:n})\sim \mathcal{Q}^{\otimes (n-1)}}\left[ \mathbb{E}_{(x, y_1)\sim \mathcal{P}} \left[ {\rm log}\,\frac{e^{f(x, y_1)}}{\frac{1}{n}\sum_{j=1}^n e^{f(x, y_j)}} \right] - 1 + 1 \right] \\
    & = \mathbb{E}_{(x, y_1)\sim \mathcal{P}, (x, y_{2:n})\sim \mathcal{Q}^{\otimes (n-1)}}\left[ {\rm log}\,\frac{e^{f(x, y_1)}}{\frac{1}{n}\sum_{j=1}^n e^{f(x, y_j)}}\right].
\end{split}
\end{equation*}
The first line comes from the fact that $D_{\rm KL} \left( \mathcal{P} \,\|\, \mathcal{Q} \right)$ is a constant. The second line comes from Lemma~\ref{lemm:kl}. The third line comes from the fact that $(x, y_1)$ and $(x, y_{2:n})$ are interchangeable when they are all sampled from $\mathcal{Q}$.

To conclude, since the inequality works for all $f$, and hence 
$$
\underset{f}{\rm sup}\,\,\mathbb{E}_{(x, y_1)\sim \mathcal{P}, (x, y_{2:n})\sim \mathcal{Q}^{\otimes (n-1)}}\left[ {\rm log}\,\frac{e^{f(x, y_1)}}{\frac{1}{n}\sum_{j=1}^n e^{f(x, y_j)}}\right] \leq D_{\rm KL} \left( \mathcal{P} \,\|\, \mathcal{Q} \right).
$$
\end{proof}
\label{lemm:infonce_like}
\end{lemma}

Note that Lemma~\ref{lemm:infonce_like} does not require $n \rightarrow \infty$, which is a much more practical setting compared to the analysis made only when $n\rightarrow \infty$. And a remark is that the equality holds in Lemma~\ref{lemm:infonce_like} when $n\rightarrow \infty$.

\subsubsection{Immediate results following Lemma~\ref{lemm:kl_3}}

\begin{lemma}
\begin{equation*}
\begin{split}
   \text{CMI}(X;Y|Z)  & = \mathbb {E} _{P_Z}\left[D_{\rm KL}\,(P_{X,Y|Z} \,\|\,P_{X|Z}P_{Y|Z})\right]
    \\
    & = D_{\rm KL}\,(P_{X,Y,Z} \,\|\,P_Z P_{X|Z}P_{Y|Z})
    \\
    & = \underset{f}{\rm sup} \,\mathbb{E}_{(x, y, z)\sim P_{X,Y,Z}} [f(x,y,z)] - \mathbb{E}_{(x, y,z)\sim P_ZP_{X|Z}P_{Y|Z}} [e^{f(x,y,z)}] + 1.
\end{split}
\end{equation*}
\begin{proof}
Let $\mathcal{P}$ be $P_{X,Y,Z}$ and $\mathcal{Q}$ be ${P_Z}P_{X|Z}P_{Y|Z}$ in Lemma~\ref{lemm:kl_3}.
\end{proof}
\label{lemm:c_infonce}
\end{lemma}

\subsubsection{Showing ${\rm \text{Weak-CMI}}\,(X;Y|Z) \leq\text{CMI}(X;Y|Z)$ }

\begin{proposition}
${\rm \text{Weak-CMI}}\,(X;Y|Z) \leq\text{CMI}(X;Y|Z)$.
\begin{proof}
According to Lemma~\ref{lemm:wc_infonce},
\begin{equation*}
\begin{split}
    {\rm \text{Weak-CMI}}\,(X;Y|Z) & = \underset{f}{\rm sup} \,\mathbb{E}_{(x, y)\sim P_{X,Y}} [f(x,y)] - \mathbb{E}_{(x, y)\sim \mathbb{E}_{P_Z}\left[P_{X|Z}P_{Y|Z}\right]} [e^{f(x,y)}] + 1 \\
    & = \underset{f}{\rm sup} \,\mathbb{E}_{(x, y, z)\sim P_{X,Y,Z}} [f(x,y)] - \mathbb{E}_{(x, y,z)\sim P_ZP_{X|Z}P_{Y|Z}} [e^{f(x,y)}] + 1.
\end{split}
\end{equation*}
Let $f_1^*(x,y)$ be the function when the equality for ${\rm \text{Weak-CMI}}\,(X;Y|Z)$ holds, and let $f_2^*(x,y,z) = f_1^*(x,y)$ ($f_2^*(x,y,z)$ will not change $\forall z \sim P_Z$):
\begin{equation*}
\begin{split}
    {\rm \text{Weak-CMI}}\,(X;Y|Z) 
    & = \mathbb{E}_{(x, y, z)\sim P_{X,Y,Z}} [f_1^*(x,y)] - \mathbb{E}_{(x, y,z)\sim P_ZP_{X|Z}P_{Y|Z}} [e^{f_1^*(x,y)}] + 1 \\
    & = \mathbb{E}_{(x, y, z)\sim P_{X,Y,Z}} [f_2^*(x,y,z)] - \mathbb{E}_{(x, y,z)\sim P_ZP_{X|Z}P_{Y|Z}} [e^{f_2^*(x,y,z)}] + 1.
\end{split}
\end{equation*}
Comparing the equation above to Lemma \ref{lemm:c_infonce},

\begin{equation*}
\begin{split}
   \text{CMI}(X;Y|Z) & = \underset{f}{\rm sup} \,\mathbb{E}_{(x, y, z)\sim P_{X,Y,Z}} [f(x,y,z)] - \mathbb{E}_{(x, y,z)\sim P_ZP_{X|Z}P_{Y|Z}} [e^{f(x,y,z)}] + 1,
\end{split}
\end{equation*}
we conclude ${\rm \text{Weak-CMI}}\,(X;Y|Z) \leq\text{CMI}(X;Y|Z)$.
\end{proof}
\label{prop:wc_leq_c_mi}
\end{proposition}

\subsection{Proof of a tighter bound of $\text{CMI}(X;Y|Z)$}
Next, we show a bound of $\text{CMI}(X;Y|Z)$ which is tighter than the proposed CCL. We term this bound as Tight-CCL.
\begin{proposition}[A tighter bound of $\text{CMI}(X;Y|Z)$]
\begin{equation*}
\small
\begin{split}
\text{Tight-CCL} &:=\underset{f}{\rm sup}\,\,\mathbb{E}_{z\sim P_Z}\left[\mathbb{E}_{(x_i, y_i)\sim {P_{X,Y|z}}^{\otimes n}}\left[ {\rm log}\,\frac{e^{f(x_i, y_i, z)}}{\frac{1}{n}\sum_{j=1}^n e^{f(x_i, y_j, z)}}\right]\right] \\
& \leq \mathbb {E} _{P_Z}\left[D_{\rm KL}\,(P_{X,Y|Z} \,\|\,P_{X|Z}P_{Y|Z})\right] =\text{CMI}(X;Y|Z),
\end{split}
\end{equation*}
\begin{proof}
Given a $z\sim P_Z$, we let $\mathcal{P} = P_{X,Y|Z=z}$ and  $\mathcal{Q} = P_{X|Z=z}P_{Y|Z=z}$. Then, 
\begin{equation*}
\begin{split}
\mathbb{E}_{(x, y_1)\sim \mathcal{P}, (x, y_{2:n})\sim \mathcal{Q}^{\otimes (n-1)}}\left[ {\rm log}\,\frac{e^{f(x, y_1, z)}}{\frac{1}{n}\sum_{j=1}^n e^{f(x, y_j, z)}}\right] & = \mathbb{E}_{(x_i, y_i)\sim {P_{X,Y|z}}^{\otimes n}}\left[ {\rm log}\,\frac{e^{f(x_i, y_i, z)}}{\frac{1}{n}\sum_{j=1}^n e^{f(x_i, y_j, z)}}\right].
\end{split}
\end{equation*}
The only variables in the above equation are $X$ and $Y$ with $Z$ being fixed at $z$, and hence the following can be obtained via Lemma~\ref{lemm:infonce_like}:
$$
 \mathbb{E}_{(x_i, y_i)\sim {P_{X,Y|z}}^{\otimes n}}\left[ {\rm log}\,\frac{e^{f(x_i, y_i, z)}}{\frac{1}{n}\sum_{j=1}^n e^{f(x_i, y_j, z)}}\right] \leq  D_{\rm KL}\,(\mathcal{P} \,\|\,\mathcal{Q}) = D_{\rm KL}\,(P_{X,Y|Z=z} \,\|\,P_{X|Z=z}P_{Y|Z=z}).
$$
The above inequality works for any function $f(\cdot, \cdot, \cdot)$ and any $z\sim P_Z$, and hence
$$
\underset{f}{\rm sup}\,\,\mathbb{E}_{z\sim P_Z}\left[\mathbb{E}_{(x_i, y_i)\sim {P_{X,Y|z}}^{\otimes n}}\left[ {\rm log}\,\frac{e^{f(x_i, y_i, z)}}{\frac{1}{n}\sum_{j=1}^n e^{f(x_i, y_j, z)}}\right]\right] \leq \mathbb {E} _{P_Z}\left[D_{\rm KL}\,(P_{X,Y|Z} \,\|\,P_{X|Z}P_{Y|Z})\right]. 
$$
\end{proof}

\label{prop:c_infonce}
\end{proposition}

We discuss the similarities and differences between CCL and Tight-CCL. Both are lower bounds of conditional mutual information $\text{CMI}(X;Y|Z)$, and both share formulations similar to InfoNCE \citep{oord2018representation}. The differences are that the scoring function $f(x, y, z)$ of Tight-CCL takes $z$ as input, while CCL does not. Taking $z$ as input makes Tight-CCL a tighter bound than CCL, which we show in Proposition \ref{prop:wc_leq_c_infonce}. The reason we do not take $z$ as an input in the CCL is because the sensitive attribute $z$ in our setup is mostly binary, carries little information, and empirically Tight-CCL performs very similar to the proposed CCL (see Section \ref{sec:exp_appendix}). CCL, on the other hand, has a simpler formulation and is easier to adapt to existing contrastive frameworks.

\subsection{Proof of Equation (4) in the Main Text}

\begin{proposition}[Conditional Contrastive Learning (CCL), restating Equation (4) in the main text]
\begin{equation*}
\small
\begin{split}
\text{CCL}&:=\underset{f}{\rm sup}\,\,\mathbb{E}_{z\sim P_Z}\left[\mathbb{E}_{(x_i, y_i)\sim {P_{X,Y|z}}^{\otimes n}}\left[ {\rm log}\,\frac{e^{f(x_i, y_i)}}{\frac{1}{n}\sum_{j=1}^n e^{f(x_i, y_j)}}\right]\right] \\
& \leq D_{\rm KL}\,\left(P_{X,Y} \,\|\,\mathbb{E}_{P_Z}\left[P_{X|Z}P_{Y|Z}\right] \right) = {\rm \text{Weak-CMI}}\,(X;Y|Z) \leq\text{CMI}(X;Y|Z).
\end{split}
\end{equation*}
\begin{proof}
By defining $\mathcal{P} = P_{X,Y}$ and $\mathcal{Q} = \mathbb{E}_{P_Z}\left[P_{X|Z}P_{Y|Z}\right]$, we have 
$$
\mathbb{E}_{(x, y_1)\sim \mathcal{P}, (x, y_{2:n})\sim \mathcal{Q}^{\otimes (n-1)}}\left[ {\rm log}\,\frac{e^{f(x, y_1)}}{\frac{1}{n}\sum_{j=1}^n e^{f(x, y_j)}}\right] = \mathbb{E}_{z\sim P_Z}\left[\mathbb{E}_{(x_i, y_i)\sim {P_{X,Y|z}}^{\otimes n}}\left[ {\rm log}\,\frac{e^{f(x_i, y_i)}}{\frac{1}{n}\sum_{j=1}^n e^{f(x_i, y_j)}}\right]\right].
$$
Via Lemma~\ref{lemm:infonce_like}, we have
$$
\underset{f}{\rm sup}\,\,\mathbb{E}_{z\sim P_Z}\left[\mathbb{E}_{(x_i, y_i)\sim {P_{X,Y|z}}^{\otimes n}}\left[ {\rm log}\,\frac{e^{f(x_i, y_i)}}{\frac{1}{n}\sum_{j=1}^n e^{f(x_i, y_j)}}\right]\right] \leq D_{\rm KL}\,\left(P_{X,Y} \,\|\,\mathbb{E}_{P_Z}\left[P_{X|Z}P_{Y|Z}\right] \right).
$$
Combing with Proposition~\ref{prop:wc_leq_c_mi} that ${\rm \text{Weak-CMI}}\,(X;Y|Z) \leq\text{CMI}(X;Y|Z)$, we conclude the proof.
\end{proof}
\label{prop:wc_infonce}
\end{proposition}

\subsection{Showing CCL is a lower bound of Tight-CCL}
\begin{proposition}
\begin{equation*}
\begin{split}
    \text{CCL}&:=\underset{f}{\rm sup}\,\,\mathbb{E}_{z\sim P_Z}\left[\mathbb{E}_{(x_i, y_i)\sim {P_{X,Y|z}}^{\otimes n}}\left[ {\rm log}\,\frac{e^{f(x_i, y_i)}}{\frac{1}{n}\sum_{j=1}^n e^{f(x_i, y_j)}}\right]\right] \\
    \leq \quad \quad \text{Tight-CCL} &:=\underset{f}{\rm sup}\,\,\mathbb{E}_{z\sim P_Z}\left[\mathbb{E}_{(x_i, y_i)\sim {P_{X,Y|z}}^{\otimes n}}\left[ {\rm log}\,\frac{e^{f(x_i, y_i, z)}}{\frac{1}{n}\sum_{j=1}^n e^{f(x_i, y_j, z)}}\right]\right].
\end{split}
\end{equation*}
\begin{proof}
Let $f_1^*(x,y)$ be the function when the equality holds in CCL, and let $f_2^*(x,y,z)=f_1^*(x,y)$ \Big( $f_2^*(x,y,z)$ will not change $\forall z \sim P_Z$ \Big):
$$
\text{CCL}:=\mathbb{E}_{z\sim P_Z}\left[\mathbb{E}_{(x_i, y_i)\sim {P_{X,Y|z}}^{\otimes n}}\left[ {\rm log}\,\frac{e^{f_2^*(x_i, y_i,z)}}{\frac{1}{n}\sum_{j=1}^n e^{f_2^*(x_i, y_j,z)}}\right]\right]. 
$$
Since the equality holds with the supreme function in Tight-CCL, and hence 
$$
\text{CCL} \leq \text{Tight-CCL}.
$$
\end{proof}
\label{prop:wc_leq_c_infonce}
\end{proposition}
\section{Experimental Details}
\label{sec:exp_appendix}
\subsection{Code}

The code for this project will be updated soon.

\subsection{Fairness Tabular Dataset Details}

\textbf{UCI Adult} \citep{Dua:2019} focuses on predicting income of a person exceeds fifty thousand per year based on census data. It has a total of $48,842$ samples, with a pre-determined training split of $32,561$ samples and a test split of $16,281$ samples. We choose the gender attribute as the sensitive attribute. It has the CC0: Public Domain License.

\textbf{UCI German} \citep{Dua:2019} focuses on predicting whether a person has good credit or not based on a set of attributes. It has a total of $10,00$ samples. We follow the split in ~\citet{ruoss2020learning}, where $80\%$ of samples are drawn randomly and used as the training set and $20\%$ samples are drawn randomly and used as the test set. We choose the age attribute as the sensitive attribute, which is determined by whether the individual’s age exceeds a threshold. It has the Database Contents License v1.0. 

\textbf{UCI Crime}: The Communities and Crime dataset \citep{Dua:2019} contains data including socioeconomic, law enforcement, and
crime information for US communities. Specifically, it focuses on predicting whether a specific community is above or below the median number of violent crimes per population. It has $1,994$ samples. We follow the split in ~\citet{ruoss2020learning}, where $80\%$ of samples are drawn randomly and used as the training set and $20\%$ samples are drawn randomly and used as the test set. We choose the race attribute as the sensitive attribute, which is determined by whether the individual has race white. It has the Database Contents License v1.0. 

\textbf{COMPAS}: The Recidivism Risk COMPAS Score dataset \citep{angwin2016machine} contains a variety of demographic and crime information collected on the use of the COMPAS risk assessment tool in Broward County, Florida Angwin. It focuses on predicting recidivism (whether a criminal will reoffend or not) in the USA. It has $5,728$ samples. We follow the split in ~\citet{ruoss2020learning}, where $80\%$ of samples are drawn randomly and used as the training set and $20\%$ samples are drawn randomly and used as the test set. We choose the predefined binary race attribute as the sensitive attribute. It has the Database Contents License v1.0.

\textbf{Law School}: The Law School dataset is from the Law School Admission Study \citep{wightman1998lsac}. It has application records for 25 different law schools. It focuses on predicting whether a student passes the law school bar exam. It has $36,022$ samples. We follow the split in ~\citet{ruoss2020learning}, where $80\%$ of samples are drawn randomly and used as the training set and $20\%$ samples are drawn randomly and used as the test set. We choose the race attribute as the sensitive attribute, which is determined by whether the individual has race white. It has the Database Contents License v1.0.

\paragraph{Dataset pre-processing}: We perform the following types of preprocessing on all five fairness datasets: first, we standardize each numerical feature of the data to zero mean and unit variance. Next, we use one-hot encoding scheme for categorical features. Then, we drop rows
and columns with missing values, and lastly we split into train, test and validation sets. For the contrastive pre-training, we augment each data sample using two noise vectors sampled from an isotropic Guassian distribution, where the variance is a hyper-parameter. For the supervised fine-tuning, all downstream classification tasks are binary prediction tasks.

\paragraph{Personal identifiable information}: Personally identifiable information is not available in all five datasets, because the authors of the datasets explicitly remove personal information when creating the datasets.

\subsection{Vision Dataset Details}
\textbf{CelebA} \citep{liu2015faceattributes} is a human facial recognition dataset that contains more than $200,000$ images of celebrity faces, where each facial image is annotated with 40 human-labeled binary attributes, including gender. Among the attributes,
we select attractive, smile, and wavy hair and use them to form three separate binary classification tasks. The sensitive attribute is gender. The license of CelebA dataset claims that it is available for noncommercial research purposes only.

\textbf{UTKFace} \citep{zhang2017age} is a human facial recognition dataset that contains more than $20,000$ images of human faces in a variety of age groups and races, where each facial image is annotated with three human-labeled binary attributes, including age, gender, and ethnicity. Among the attributes, we select age as the binary classification task (if the age of the individual is above a threshold). The sensitive attribute is the race attribute. The license of UTKFace dataset claims that it is available for noncommercial research purposes only.

\paragraph{Dataset pre-processing}: For CelebA, we directly use the pre-defined training and test sets from the PyTorch data loader for CelebA. For UTKFace, we use a random $20\%$ of all samples as the test set. The data augmentation details of both datasets will be included in Section \ref{sec:training}. 

\paragraph{Personal identifiable information}: Personally identifiable information is not available in the UTKFace data set, because the authors of the data sets explicitly remove personal information when creating the data set. For the CelebA dataset, each person has an ID, but the identity is not explicitly revealed (although users can infer the identities of some celebrities). Both datasets contain the annotated information, such as age, gender, and other facial attributes of the individuals in the images.

\subsection{Fairness Tabular Dataset Training Details and Results}
\label{sec:fair_result_appendix}
We follow the implementation from \citep{ruoss2020learning}. We use a three-layer neural network with hidden dimension 100 as the encoder and a linear layer as the fine-tuning network. We train $100$ epochs and report the result. For pre-training, we use the Adam \citep{kingma2014adam} optimizer, with a batch size of $256$, a learning rate of $0.001$, and a weight decay of $0.01$. For fine-tuning, we use the same optimizer, batch size, weight decay, but a slightly larger learning rate $0.005$.
\vspace{-4mm}
\begin{table}[H]
\centering
\caption{Accuracies and fairness results on five fairness datasets with confidence intervals. CCL has better downstream accuracy than existing unsupervised and self-supervised baselines in four datasets, and exhibits better fairness measurements in most cases.}
\vspace{-2mm}

\begin{adjustbox}{center,width=0.85\linewidth}
\begin{tabular}{lllcccc}
\toprule

&&Model &Accuracy (\%) $(\uparrow)$ & $\Delta_{DP}\,(\downarrow)$ & $\Delta_{EO}\,(\downarrow)$ & $\Delta_{{EO}_{PP}}\,(\downarrow)$      \\ 
\midrule

\multicolumn{2}{c}{\parbox[t]{2mm}{\multirow{9}{*}{\rotatebox[origin=c]{90}{ADULT}}}} & \bf Unsupervised\\
&&--  LAFTR \citep{madras2018learning} & 84.0 & 0.163 & \bf 0.030 & \bf 0.026 \\
&&-- \citet{ragonesi2021learning} & 85.0 & - & \bf 0.030 & -  \\
&&-- DTM \citep{lee2020maximal} & 71.6 & -& 0.050 & -\\
&&-- FNF \citep{balunovic2021fair} & 80.0 & \bf 0.110 & - & - \\
&&\bf Self-Supervised & \\
&&-- FairMixRep \citep{chakraborty2020fairmixrep}  & 85.0 & 0.172 & - & - \\
&&-- SimCLR \citep{chen2020simple} & $83.1 \pm 0.48$ & $0.210 \pm 0.04$ & $0.410 \pm 0.05$ & $0.320 \pm 0.04$\\
&&-- \bf CCL (Ours) & $\textbf{85.4} \pm 0.53$ & $\textbf{0.110} \pm 0.02$ & $0.070 \pm 0.01$  & $0.090 \pm 0.01$ \\
&&-- \bf Tight-CCL (Ours) & $\textbf{85.3} \pm 0.48$ & $\textbf{0.108} \pm 0.03$ & $0.068 \pm 0.01$  & $0.093 \pm 0.01$ \\

\midrule
\multicolumn{2}{c}{\parbox[t]{2mm}{\multirow{6}{*}{\rotatebox[origin=c]{90}{COMPAS}}}} &\bf Unsupervised \\
&&-- DTM \citep{lee2020maximal} & 66.0 & - & 0.200 & - \\
&&-- FNF \citep{balunovic2021fair} & 65.0 & 0.240 & - & - \\
&&\bf Self-Supervised & \\
&&-- SimCLR \citep{chen2020simple}  & $\textbf{71.2} \pm 0.33$ & $0.103 \pm 0.01 $ & $0.227 \pm 0.06$ & $0.134 \pm 0.04$\\
&&-- \bf CCL (Ours) & $\textbf{71.0} \pm 0.25$& $\textbf{0.080} \pm 0.01$ & $\textbf{0.132} \pm 0.03$  & $\textbf{0.081} \pm 0.01$  \\
&&-- \bf Tight-CCL (Ours) & $\textbf{70.8} \pm 0.28$ & $\textbf{0.090} \pm 0.01$ & $\textbf{0.142} \pm 0.02$  & $\textbf{0.101} \pm 0.02$ \\


\midrule
\multicolumn{2}{c}{\parbox[t]{2mm}{\multirow{6}{*}{\rotatebox[origin=c]{90}{CRIME}}}}&\bf Unsupervised&\\

&&-- LCIFR \citep{ruoss2020learning} & \bf 84.4 & 0.443 & 0.314 & 0.212 \\
&&-- FNF \citep{balunovic2021fair} & 82.5 & 0.540 & - & - \\
&&\bf Self-Supervised\\
&&-- SimCLR \citep{chen2020simple} & $82.1 \pm 0.32$ & $0.502 \pm 0.08$ & $0.530 \pm 0.02$ & $0.383 \pm 0.01$ \\
&&-- \bf CCL (Ours) & $82.6 \pm 0.24$ & $\textbf{0.211} \pm 0.02$ & $\textbf{0.224} \pm 0.02$  & $\textbf{0.183} \pm 0.01$ \\
&&-- \bf Tight-CCL (Ours) & $82.5 \pm 0.30$ & $\textbf{0.208} \pm 0.02$ & $\textbf{0.222} \pm 0.02$  & $\textbf{0.181} \pm 0.02$ \\

\midrule
\multicolumn{2}{c}{\parbox[t]{2mm}{\multirow{7}{*}{\rotatebox[origin=c]{90}{GERMAN}}}}& \bf Unsupervised\\
&&-- LCIFR \citep{ruoss2020learning} & 73.1 & 0.102 & 0.080 & 0.063 \\

&&-- \citet{ragonesi2021learning} & 74.0 & - & \bf 0.060 & -  \\
&&\bf Self-Supervised\\
&&-- FairMixRep \citep{chakraborty2020fairmixrep}  & 71.8 & 0.089 & - & - \\

&&-- SimCLR \citep{chen2020simple}   & $72.5 \pm 0.11$ & $0.250 \pm 0.05$ & $0.382 \pm 0.06$ & $0.195 \pm 0.04$ \\
&&-- \bf CCL (Ours) & $\textbf{74.3} \pm 0.28$ & $\textbf{0.083} \pm 0.01$ & $ 0.128 \pm 0.03$  & $ \textbf{0.062} \pm 0.01$ \\
&&-- \bf Tight-CCL (Ours) & $\textbf{74.4} \pm 0.25$ & $\textbf{0.085} \pm 0.01$ & $0.126 \pm 0.02$  & $\textbf{0.062} \pm 0.01$ \\

\midrule
\parbox[t]{0.1mm}{\multirow{6}{*}{\rotatebox[origin=c]{90}{LAW}}} & \parbox[t]{2mm}{\multirow{6}{*}{\rotatebox[origin=c]{90}{SCHOOL}}} &
\bf Unsupervised\\ 
&&-- LCIFR \citep{ruoss2020learning} & 84.4 & 0.110 & 0.180 & 0.070 \\
&&-- FNF \citep{balunovic2021fair} & 84.6 & \bf 0.050 & - & - \\
&&\bf Self-Supervised\\
&&-- SimCLR \citep{chen2020simple}   & $83.6 \pm 0.54$ & $0.086 \pm 0.02$ & $0.212  \pm 0.04$ & $0.110 \pm 0.02$ \\
&&-- \bf CCL (Ours) & $\textbf{84.8} \pm 0.50$ & $\textbf{0.051} \pm 0.01$ & $\textbf{0.153} \pm 0.03$  & $\textbf{0.056} \pm 0.01$\\
&&-- \bf Tight-CCL (Ours) & \bf $\textbf{84.5} \pm 0.44$ & $\textbf{0.050} \pm 0.01$ & $\textbf{0.150} \pm 0.02$  & $\textbf{0.055} \pm 0.01$\\


\bottomrule
\end{tabular}
\end{adjustbox}
\label{tab:fair_std}

\end{table}
\vspace{-8mm}

\paragraph{Results.} We include the results in Table \ref{tab:fair_std}. All entries with $-$ indicate that the corresponding metrics are not reported in the original papers. The following results are read off from the figures in the paper: LAFTR \citep{madras2018learning}, DTM \citep{lee2020maximal}, and FNF \cite{balunovic2021fair}. We include the confidence intervals of the results, and bold the entries that have overlapping confidence intervals with the best performing entries in that dataset. SimCLR is a re-implementation of \citet{chen2020simple} on the new datasets. Tight-CCL represents a tighter bound of conditional mutual information, which is introduced and discussed in Proposition \ref{prop:c_infonce}. From the results, we can conclude that CCL outperforms all baselines on eleven out of the eighteen fairness metrics. Also, CCL outperforms all baselines on downstream accuracy on four out of five datasets. We note that Tight-CCL performs very close to CCL, sometimes better than CCL in terms of fairness metrics. This may be due to that the Tight-CCL is a tighter bound of conditional mutual information, and optimizing Tight-CCL leads to a representation closer to conditional mutual information maximization. Because conditional mutual information explicitly excludes information from the sensitive attribute $Z$, Tight-CCL is able to remove slightly more effect from the sensitive attribute than CCL. We use CCL in the main text as it has a simpler formulation and is easier to adapt to existing contrastive frameworks.

\paragraph{Computational resource} We perform all experiments on a single GeForce
RTX 2080 Ti GPU and a 32-core Intel CPU processor. Training $100$ epochs in different datasets vary based on the size of the dataset, but the overall training time of $100$ epochs on one dataset is below an hour.

\subsection{Vision Dataset Training Details and Results}

We follow the implementation from \citep{chen2020simple}. We use a ResNet-50 as the encoder and a two-layer network with hidden dimension $512$ as the fine-tuning network. We train $100$ epochs and report the result. For the contrastive pre-training, we use the Adam \citep{kingma2014adam} optimizer, with a batch size of $256$, a learning rate of $0.0003$, and a weight decay of $10^{-6}$. For the supervised fine-tuning, we use the same optimizer, batch size, weight decay, but a slightly larger learning rate $0.001$.

\begin{table}[H]
\centering
\caption{Accuracies and fairness results on two vision datasets on four prediction tasks with confidence intervals. Best results are bold. CCL has better downstream accuracy in all four tasks, and exhibits better or close-to-the-best fairness measurements in 11 out of 12 results.}

\begin{adjustbox}{center,width=0.85\linewidth}
\begin{tabular}{lllcccc}
\toprule

&&Model &Accuracy (\%) $(\uparrow)$ & $\Delta_{DP}\,(\downarrow)$ & $\Delta_{EO}\,(\downarrow)$ & $\Delta_{{EO}_{PP}}\,(\downarrow)$      \\ 
\midrule

\parbox[t]{0.1mm}{\multirow{7}{*}{\rotatebox[origin=c]{90}{CELEBA}}} & \parbox[t]{2mm}{\multirow{7}{*}{\rotatebox[origin=c]{90}{ATTRACTIVE}}} &
\bf Unsupervised\\ 
&&-- MFD \citep{jung2021fair} & 80.2 & - & \bf 0.050 & - \\
&&-- \citet{balunovic2021fair} & 79.4 & - & 0.238 & - \\
&&-- \citet{morales2020sensitivenets} & 77.7 & - & 0.070 & - \\
&&\bf Self-Supervised\\
&&-- SimCLR \citep{chen2020simple}   & $\textbf{81.7} \pm 0.32$ & $0.277 \pm 0.04$ & $0.212 \pm 0.03$ & $0.110 \pm 0.01$ \\
&&-- \bf CCL (Ours) & $\textbf{82.1} \pm 0.24$  & $\textbf{0.202} \pm 0.03$ & $0.101 \pm 0.01$  & $\textbf{0.048} \pm 0.01$\\
&&-- \bf Tight-CCL (Ours) & $\textbf{81.9} \pm 0.33$  & $\textbf{0.200} \pm 0.02$ & $0.106 \pm 0.02$  & $\textbf{0.052} \pm 0.01$\\

\midrule
\parbox[t]{0.1mm}{\multirow{6}{*}{\rotatebox[origin=c]{90}{CELEBA}}} & \parbox[t]{2mm}{\multirow{6}{*}{\rotatebox[origin=c]{90}{WAVY HAIR}}} &
\bf Unsupervised\\ 
&&-- FactorVAE \citep{kim2018disentangling} & 64.5 & - & 0.388 & 0.288 \\
&&-- FFVAE \citep{creager2019flexibly} & 61.0 & - & 0.211 & 0.154 \\
&&\bf Self-Supervised\\
&&-- SimCLR \citep{chen2020simple}   & $\textbf{67.7} \pm 0.76$ & $0.403 \pm 0.05$ & $0.355 \pm 0.04$ & $0.210 \pm 0.02$ \\
&&-- \bf CCL (Ours) & $\textbf{67.7} \pm 0.69$ & $\textbf{0.202} \pm 0.02$ & $\textbf{0.189} \pm 0.02$  & $\textbf{0.102} \pm 0.01$\\
&&-- \bf Tight-CCL (Ours) & $\textbf{67.8} \pm 0.44$ & $\textbf{0.198} \pm 0.03$ & $\textbf{0.172} \pm 0.02$  & $\textbf{0.093} \pm 0.01$\\

\midrule
\parbox[t]{0.1mm}{\multirow{5}{*}{\rotatebox[origin=c]{90}{CELEBA}}} & \parbox[t]{2mm}{\multirow{5}{*}{\rotatebox[origin=c]{90}{SMILE}}} &
\bf Unsupervised\\ 
&&-- \citet{morales2020sensitivenets} & 88.4 & - & \bf 0.060 & - \\

&&\bf Self-Supervised\\
&&-- SimCLR \citep{chen2020simple}   & $\textbf{89.3} \pm 0.33$ & $0.102 \pm 0.01$ & $0.142 \pm 0.01$ & $0.078 \pm 0.01$ \\
&&-- \bf CCL (Ours) & $\textbf{89.7} \pm 0.25$ & $\textbf{0.086} \pm 0.01$ & $\textbf{0.060} \pm 0.01$  & $\textbf{0.053} \pm 0.01$\\
&&-- \bf Tight-CCL (Ours) & $\textbf{89.5} \pm 0.27$ & $\textbf{0.084} \pm 0.01$ & $\textbf{0.060} \pm 0.01$  & $\textbf{0.056} \pm 0.01$\\

\midrule
\parbox[t]{0.1mm}{\multirow{6}{*}{\rotatebox[origin=c]{90}{UTKFACE}}} & \parbox[t]{2mm}{\multirow{6}{*}{\rotatebox[origin=c]{90}{GENDER}}} &
\bf Unsupervised\\ 
&&-- AD \citep{zhang2018mitigating} & 74.7 & - & 0.204 & - \\
&&-- MFD \citep{jung2021fair} & 74.7 & - & 0.178 & - \\
&&\bf Self-Supervised\\
&&-- SimCLR \citep{chen2020simple}   & $78.0 \pm 0.25$ & $0.335 \pm 0.03$ & $0.421 \pm 0.04$ & $0.287 \pm 0.03$ \\
&&-- \bf CCL (Ours) & $\textbf{78.5} \pm 0.22$ & $\textbf{0.191} \pm 0.02$ & $\textbf{0.156} \pm 0.02$  & $\textbf{0.089} \pm 0.01$\\
&&-- \bf Tight-CCL (Ours) & $\textbf{78.3} \pm 0.15$ & $\textbf{0.188} \pm 0.02$ & $\textbf{0.159} \pm 0.02$  & $\textbf{0.110} \pm 0.02$\\

\bottomrule
\end{tabular}
\end{adjustbox}
\label{tab:vision_std}

\end{table}

\paragraph{Results.} We include the results in Table \ref{tab:vision_std}. All entries with $-$ indicate that the corresponding metrics are not reported in the original papers. We include the confidence intervals of the results, and bold the entries that have overlapping confidence intervals with the best performing entries in that dataset. Similar to our observation in Section \ref{sec:fair_result_appendix}, from the results we can conclude that CCL outperforms all baselines on eleven out of the twelve fairness metrics. Also, CCL outperforms all baselines on downstream accuracy on all four tasks.  Tight-CCL also performs very close to CCL, although some downstream task performances of Tight-CCL is slightly worse than that of CCL.

\paragraph{Computational resource} We perform all experiments on a single GeForce
RTX 2080 Ti GPU and a 32-core Intel CPU processor. Training $100$ epochs on CelebA or UTKFace takes approximately $20-24$ hours, depending on the server's condition.
\label{sec:training}

\end{document}